\pdfoutput=1

\documentclass[11pt]{article}

\usepackage[final]{acl}

\usepackage{times}
\usepackage{latexsym}
\usepackage[final]{changes} 
\usepackage[T1]{fontenc}

\usepackage[utf8]{inputenc}
\usepackage{makecell}
\usepackage{multirow} 
\usepackage{microtype}
\usepackage{tcolorbox}
\usepackage{inconsolata}

\usepackage{graphicx}

\title{Step-Opt: Boosting Optimization Modeling in LLMs through Iterative Data Synthesis and Structured Validation}

\author{
 \textbf{Yang Wu\textsuperscript{1,2}},
 \textbf{Yifan Zhang\textsuperscript{1,2,4}$^{*}$},
 \textbf{Yurong Wu\textsuperscript{2}},
 \textbf{Yuran Wang\textsuperscript{3}},
 \textbf{Junkai Zhang\textsuperscript{1,2}},
 \textbf{Jian Cheng \textsuperscript{1,2,5}}
\\
\\
 \textsuperscript{1}C$^2$DL, Institute of Automation, Chinese Academy of Sciences \\
 \textsuperscript{2}School of Artificial Intelligence, University of Chinese Academy of Sciences, Beijing \\
   \textsuperscript{3}Dalian Minzu University
 \textsuperscript{4}University of Chinese Academy of Sciences, Nanjing
 \textsuperscript{5}AIRIA
 \\
{
   \textbf{Correspondence:} {yfzhang@nlpr.ia.ac.cn}
}
}

\begin{document}
\maketitle
\begin{abstract}
Large Language Models (LLMs) have revolutionized various domains but encounter substantial challenges in tackling optimization modeling tasks for Operations Research (OR),  particularly when dealing with complex problem. In this work, we propose Step-Opt-Instruct, a framework that augments existing datasets and generates high-quality fine-tuning data tailored to optimization modeling. Step-Opt-Instruct employs iterative problem generation to systematically increase problem complexity and stepwise validation to rigorously verify data, preventing error propagation and ensuring the quality of the generated dataset. Leveraging this framework, we fine-tune open-source LLMs, including LLaMA-3-8B and Mistral-7B, to develop Step-Opt—a model that achieves state-of-the-art performance on benchmarks such as NL4OPT, MAMO, and IndustryOR. Extensive experiments demonstrate the superior performance of Step-Opt, especially in addressing complex OR tasks, with a notable 17.01\% improvement in micro average accuracy on difficult problems. These findings highlight the effectiveness of combining structured validation with gradual problem refinement to advance the automation of decision-making processes using LLMs. 
The code and dataset are available at \href{https://github.com/samwu-learn/Step}{https://github.com/samwu-learn/Step}.
\end{abstract}

\section{Introduction} 
\label{label:intro}
Operations Research (OR) is a valuable discipline for addressing complex decision-making problems, widely applied in fields such as economics, engineering, and computer science \citep{or_example1, or_example3, or_example2}. Effective implementation of OR involves two essential steps: modeling real-world problems and solving them. Despite significant advancements in solution techniques and the development of more efficient solvers, constructing appropriate models remains a challenge. Such a task requires formulating natural language descriptions into mathematical models, which is labor-intensive and demands domain-specific expertise as well as a deep understanding of modeling methodologies.  These requirements greatly restrict the broader application of OR, particularly in real-world scenarios.

Recent developments in Large Language Models (LLMs) have enhanced the feasibility of automating optimization modeling. Approaches like Chain-of-Experts (CoE) \citep{chain-of-expert} and OptiMUS \citep{optimus} employ well-crafted prompts and multi-agent systems to enhance the construction of optimization models and corresponding programs. However, these approaches rely on general-purpose LLMs, which, though powerful, are not specifically tailored for OR, limiting their effectiveness in addressing specialized challenges. Additionally, the need to upload sensitive data poses additional privacy concerns. 
In response, ORLM \citep{orlm} presents an alternative by fine-tuning open-source LLMs using a dataset of 30K examples generated from 686 industry cases. While this improves the model's performance for OR modeling, ORLM remains semi-automated, requiring significant manual post-processing to achieve satisfactory results. Moreover,  its prompt design lacks the precision needed to manage problem complexity and diversity, resulting in suboptimal outputs. Furthermore, modeling errors are not identified in real-time, allowing inaccuracies to persist and propagate. While rule-based post-processing can address minor errors, it often fails to rectify deeper logical and structural issues, further compromising data quality.

\begin{figure*}[ht]
\begin{center}
\centerline{\includegraphics[width=0.8\textwidth, trim=0cm 0cm 0cm 0cm, clip]{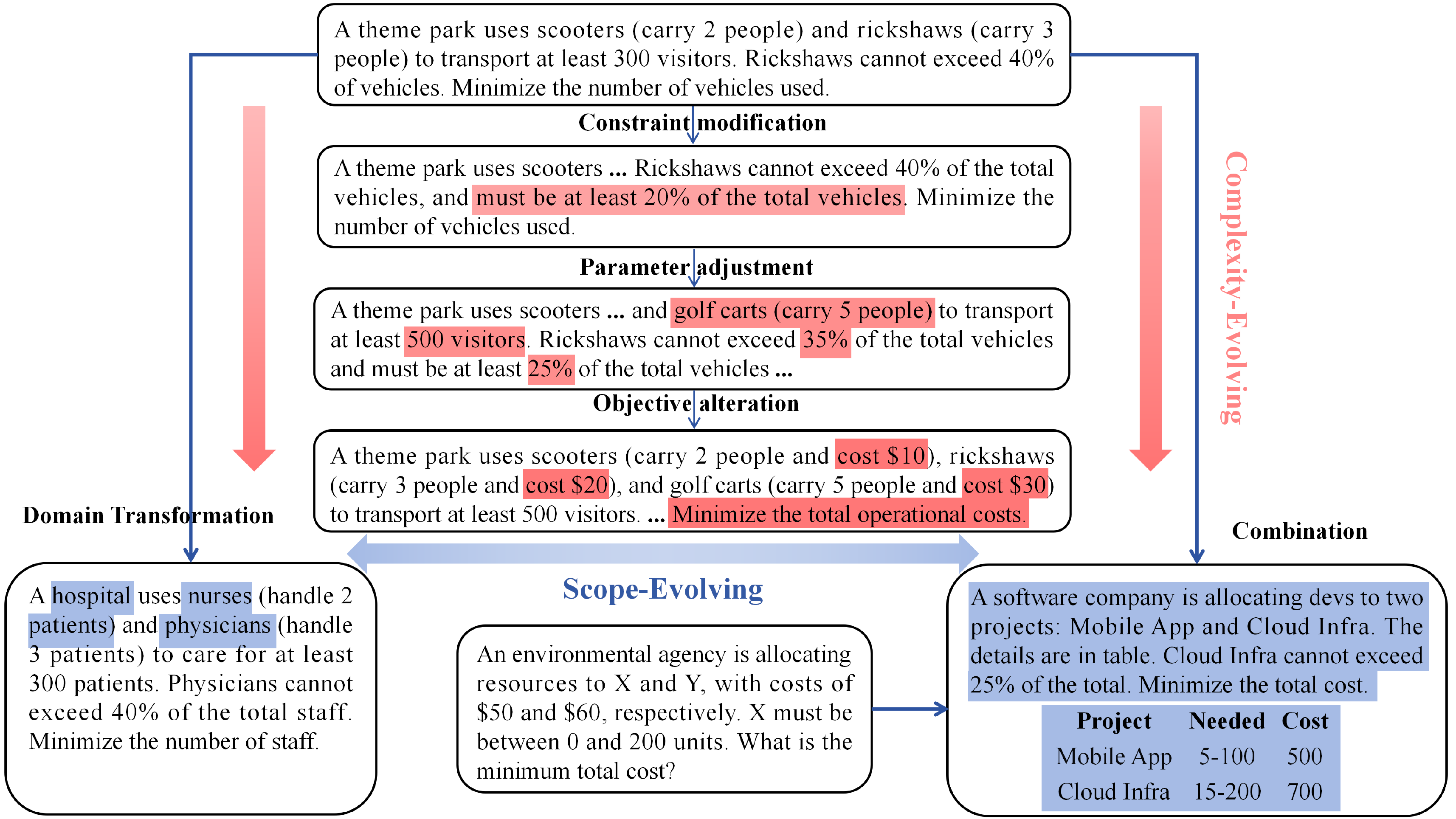}}

\caption{Examples of Iterative Problem Generation. It includes two types of methods: Complexity-Evolving, which refines problem complexity through constraint modification, parameter adjustment, and objective alteration; and Scope-Evolving, which enhances diversity via domain transformations and problem combinations. Red and blue backgrounds indicate changes introduced by Complexity- and Scope-Evolving, respectively. Repeated content is replaced with "..." for clarity.}
\label{fig:intro}
\end{center}
\vskip -0.2in
\end{figure*}

Utilizing high-quality training data is vital for improving the modeling capabilities of LLMs. However, current methods not only rely heavily on manual post-processing but also struggle to ensure data reliability. To address these limitations, we propose an approach from two primary perspectives. First, we enhance the prompt design and introduce an Iterative Problem Generation, as shown in Figure \ref{fig:intro}. This method incrementally increases the complexity and scope of the problems, allowing the dataset to retain varying levels of difficulty and breadth. This diversity plays a crucial role in improving the model's generalization capabilities, as WizardLM \citep{wizardlm} suggested. Second, we incorporate a stepwise validation mechanism that performs real-time checks throughout the generation process, effectively filtering out low-quality or erroneous data. This prevents errors from entering and propagating through the seed dataset. We refer to this framework as \textbf{Step-Opt-Instruct}. Our framework eliminates the need for post-processing, enabling fully automated generation while reducing API costs by utilizing only high-quality data for future iterations.

Step-Opt-Instruct consists of two key components: Iterative Problem Generation and Stepwise Validation Mechanism. The Iterative Problem Generation 
is specifically designed to address the unique challenges of OR-specific tasks, such as complex variable definitions and strict constraint implementation. By employing tailored methods such as Complexity-Evolving and Scope-Evolving, this approach generates a dataset enriched with enhanced complexity and diversity from the given one, facilitating the fine-tuning of LLMs to enhance their modeling ability for OR problems. As illustrated in Figure \ref{fig:intro}, Complexity-Evolving increases the complexity of the problem refining constraints, objectives, or parameters, while Scope-Evolving expands linguistic diversity and problem scope by adapting problems to new contexts or merging scenarios. This approach ensures that the generated dataset captures a wide range of complexities and provides robust coverage.

As new generated problems become increasingly complex, current LLMs often struggle to solve them accurately, resulting in errors.   If these errors remain undetected and uncorrected, they will propagate through the iterative process, ultimately affecting the quality of the generated data. To mitigate this, the stepwise validation mechanism is implemented to not only prevent errors but also guarantee the accurate application of essential modeling techniques. Problems are first validated via a description checker for completeness, followed by checks on variables, constraints, and programs. Identified issues are resolved  via feedback loops, with advanced techniques like the Big-M method verified using specially designed prompts that guide the LLM step-by-step to confirm accurate implementation.  This validation process enables the generation of reliable and high-quality datasets, which are crucial for fine-tuning LLMs and enhancing their modeling ability for OR problems.

In order to evaluate the effectiveness of Step-Opt-Instruct, we collect 260 seed cases and generate nearly 4.5K examples. This data is then applied to train LLaMA-3-8B \citep{llama3} and Mistral-7B \citep{mistral}, producing a model named Step-Opt. Furthermore, we manually review benchmarks including NL4OPT \citep{nl4opt}, MAMO \citep{mamo}, and IndustryOR \citep{orlm}, correcting a large number of examples with error labels. Experiments across these benchmarks indicate that our method outperforms existing approaches, achieving a 6.07\% improvement in the micro average and a 7.93\% enhancement in the macro average.  Notably, when focusing on more complex components, Step-Opt exhibits a more significant advantage, attaining improvements of 17.01\% and 12.26\% in micro and macro averages, respectively. This substantial lead underscores our method's capability to manage complex problems effectively.

Our contributions are as follows:\par
\textbullet  \enspace Introduction of advanced feedback mechanisms and real-time data updates, significantly reducing error propagation, \deleted{thereby} eliminating the need for extensive manual post-processing.\par
\textbullet  \enspace  Development of Step-Opt-Instruct, a novel framework specifically designed to enhance the capabilities of open-source LLMs for effectively modeling OR problems.
 \par 
\textbullet  \enspace  Proposal of the Step-Opt model, which achieves state-of-the-art performance across several benchmarks and particularly for complex problems, with additional manual corrections applied to errors in established benchmarks such as NL4OPT, MAMO, and IndustryOR.
\section{Related Work} 
\label{label:related_work}
\textbf{LLM-based Automated Modeling for OR} is an emerging field that uses LLMs to generate mathematical models for OR problems. Existing methods can be categorized into prompt-engineering and fine-tuning. Approaches like Chain-of-Thought \citep{cot} and Reflexion \citep{reflexion} improve performance but are not specialized for OR. More advanced methods, including OptiGuide \citep{optguide}, Chain-of-Experts \citep{chain-of-expert}, and OptiMUS \citep{optimus}, employ multi-agent systems with LLM to construct models but encounter difficulties with complex problems due to LLM’s limitations.  ORLM \citep{orlm}, conversely,  utilizes 
dataset generated from industry cases and GPT-4, coupled with rule-based post-processing, to fine-tune LLMs and improve outcomes. However, it lacks precise prompt 
and effective filtering mechanisms. Our framework addresses these limitations by iterative-based generation and real-time validation to control complexity and minimize errors, thereby enhancing the performance.

\textbf{Data Augmentation} improves LLM performance by generating synthetic datasets, often used when real-world data is insufficient for complex tasks\citep{self-instruct, example_of_related_work6,example_of_related_work3, example_of_related_work4,example_of_related_work2,wizardlm,example_of_related_work5}.   In operations research, data augmentation approaches like \citep{or_data_1, or_data_2} focus on synthesizing optimization problems from natural language descriptions, but with limited complexity. ORLM \citep{orlm} expands industry case datasets through modifications and rephrasings, while ReSocratic \citep{or_data_3} takes a reverse data synthesis approach, generating optimization scenarios from solutions. Among all these works, the closest to ours is Evol-Instruct \citep{wizardlm}, which uses In-depth Evolving and In-breadth Evolving to generate instruction data. However, as OR modeling presents unique challenges, we propose a stepwise validation mechanism to ensure accuracy and avoid error propagation in generated data.

\begin{figure*}[ht]
\begin{center}
\centerline{\includegraphics[width=0.85\textwidth, trim=0cm 0cm 0cm 0cm, clip]{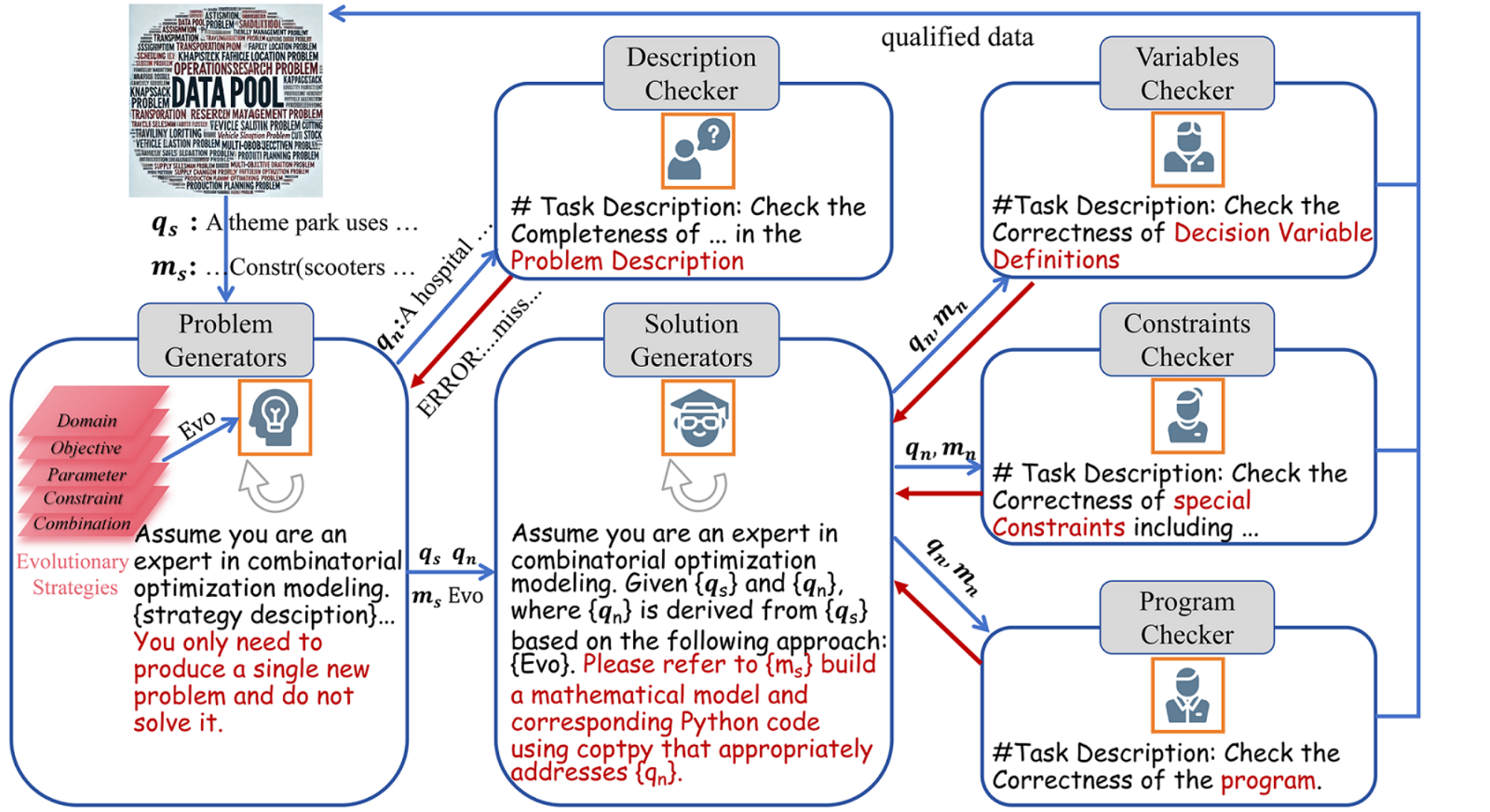}}

\caption{The framework of Step-Opt-Instruct. Each iteration begins by sampling seed data from an initial dataset. The Problem Generator uses evolving prompts to create a problem description, which is refined through feedback from the Description Checker. Once approved, the Solution Generator produces a solution, validated by the Variables, Constraints, and Program Checkers. Detected errors are fed back for revision. Only verified descriptions and solutions are added to the dataset. All components rely on tailored prompts to ensure quality. Red text indicates prompt customizations. The resulting  dataset is then used to fine-tune LLMs, improving modeling capabilities.}
\label{fig:framework}
\end{center}
\vskip -0.2in
\end{figure*}

\section{Method} 
\label{label:method}

This section outlines the proposed framework. As depicted in Figure \ref{fig:framework}, It comprises two primary components: generators and a stepwise validation mechanism. The details of the generators are provided in Sec. \ref{subsec:generator}, while the stepwise validation mechanism is detailed in Sec. \ref{subsec:checker}.

\subsection{Preliminary}
We start generation from a given initial dataset, denoted as $D = \{(q_i, m_i)\}_{i=1}^{K}$, where each instance includes a problem description $q_i$ and its associated mathematical model and program $m_i$. A valid $q_i$ must contain an objective function, constraints, and all relevant parameters with specified numerical values. The model $m_i$ implements the constraints and objective functions defined in $q_i$ and produces executable code. An example of the training data is provided in Appendix \ref{subsec:training example}. The parameter $K$ denotes the size of the initial seed dataset.

\subsection{Generators} \label{subsec:generator}

The problem generator adopts an iterative  methodology, progressively producing problems with increasing complexity and diversity. In each iteration, a seed data $(q_s, m_s)$ is randomly sampled. A specific evolving method, denoted as $f_{e}$, is then applied to create a new problem description $q_n = f_{e}(q_s)$. This process employs prompt-based LLM methods to refine problem descriptions and systematically expand their scope. These methods can be categorized into two types: \textbf{Complexity-Evolving} and \textbf{Scope-Evolving}, as detailed below.

\textbf{Complexity-Evolving} increases complexity by modifying existing conditions or introducing new elements. Considering the specific characteristics of OR problems, three approaches are included: constraint modification, objective alteration, and parameter adjustment. These incrementally raise complexity while preserving logical integrity.



Constraint modification revises existing constraints or adds new ones to enhance the problem, with the core principle being to \textit{"modify constraints based on the given problem while retaining its logical structure."} This ensures that the essential logic of the problem remains intact as complexity increases.  Similarly, objective alteration either modifies objectives or introduces new ones, with the restriction that changes cannot merely adjust coefficients. Parameter adjustment changes values or adds elements.  These approaches, while tailored to specific contexts, share the principle of preserving the underlying structure. Together, they enhance problem difficulty from various perspectives.

Nevertheless, the generated problems may become so complex that they exceed the processing capabilities of LLMs. To manage this, modifications to constraints or objectives are limited to one at a time, and parameter adjustments introduce at most one new entity. Specifically, the prompt for constraint modification is subject to strict limitations: only one constraint can be modified or added per iteration to control the growth in complexity. These restrictions not only ensure a balanced dataset with varying difficulty levels but also exclude excessively challenging examples, thereby improving the model's generalization capabilities. Prompt templates are provided in Appendix~\ref{subsec:depth}.

\textbf{Scope-Evolving} broadens topic coverage and diversity by transforming the seed example into a different domain or combining it with another example to create a novel scenario. Domain transformation transfers the basic structure of the original problem to a new domain, while preserving its logic and constraints, thereby increasing linguistic and contextual diversity. To ensure practical relevance, we define a list of reference domains. Alternatively, the combination approach merges two distinct problems to create a new one that belongs to a different domain and contains unique details. This approach introduces more substantial changes. To control complexity, the new problem is required to be of a similar length to one of the originals, maintaining manageable difficulty.  Prompt templates for Scope-Evolving are provided in Appendix~\ref{subsec:breadth}.

As the Complexity- and Scope-Evolving progress, the complexity, scope, and diversity of generated data expand, ensuring comprehensive coverage across dimensions. All approaches use two-shot examples to maintain consistency.

\textbf{Solution generator} $g$ produces a mathematical model and program $m_n$ for a valid problem $q_n$. It generates $m_n=g(q_n, q_s, m_s, f_e)$ using $q_s$, $m_s$ and evolving method $f_e$ as references. Since LLMs may struggle with complex models, we embed the instruction \textit{"ensuring the format and structure are as consistent as possible with the provided $q_s$ and $m_s$"} into the meta-prompt to enforce consistency.
\subsection{Stepwise Validation Mechanism}\label{subsec:checker}
While the aforementioned generation methods can produce descriptions and solutions, the complexity of OR problem modeling poses significant challenges for current LLMs, often causing missing parameters, ambiguous objectives, or misused advanced techniques. Without sufficient supervision and error-correction mechanisms, such issues tend to persist, gradually undermining dataset quality and negatively impacting model performance.

To address these challenges, we design a stepwise validation mechanism that checks throughout the generation process, eliminating low-quality or erroneous data to maintain dataset integrity. This mechanism comprises four checkers, each focusing on a specific aspect: description completeness, variable definition, constraint implementation, and program quality. The description checker evaluates whether the generated $q_n$ contains essential components. If any is missing, the checker provides feedback, prompting regeneration until validation succeeds or the attempt limit is reached. Only after passing this check does the solution generator produce the mathematical model and program.

Subsequently, additional checkers cross-reference $q_n$ and $m_n$ to conduct assessments. For variables, step-by-step instructions are provided, along with examples covering common types, enabling the checker to verify variable definitions.

The constraint checker ensures constraints are correctly formulated and aligned with the problem description. It follows a systematic process:  identifying constraints, then verifying their consistency with the problem’s content, similar to variable validation. While all constraints are reviewed, special attention is given to advanced techniques such as the Big-M method 
and K-way selection constraints. These serve as specialized checks, with other advanced techniques also applicable. Finally, the program checker extracts and executes the program, capturing outputs or errors, and providing feedback to the solution generator as needed.

When errors are identified in $m_n$, they are relayed back to the solution generator with the prompt: \textit{"Please regenerate the solution based on the 'Error'. Ensure that the new solution correctly addresses the problem while maintaining the format and structure, with only the necessary corrections and improvements." } The revised solution undergoes further testing until it passes all validation stages. If the retry limit is reached, the problem will be discarded. This validation process ensures both $q_n$ and $m_n$ are error-free. Only data that pass all assessments are integrated into the dataset $D$ for future iterations. This minimizes errors within $D$, thereby preventing the propagation of inaccuracies in future generations and safeguarding overall dataset quality. Details of the checkers and regeneration are provided in Appendix \ref{others}

\section{Experiment} 
\label{label:experiment}

\begin{table*}[ht]
\caption{Performance comparison of methods. Values
marked with a $^{*}$ are directly copied from original papers.}
\label{tab:performance-table}
\vskip 0.1in
\begin{center}
\small 
\setlength{\tabcolsep}{4pt} 
\begin{tabular}{llccccccc}
\hline
&\multirow{2}{*}{\textbf{Method}} & \multirow{2}{*}{\textbf{NL4OPT}} & \multirow{2}{*}{\textbf{MAMO} \vspace{0.2cm}} & \multirow{2}{*}{\textbf{MAMO} \vspace{0.2cm}} & \multirow{2}{*}{\textbf{IndustryOR}} & \multirow{2}{*}{\textbf{Micro Avg}} & \multirow{2}{*}{\textbf{Macro Avg}} \\
& &  & \textbf{EasyLP} & \textbf{ComplexLP} &  &  &  \\
\hline
\textit{PLMs} &tag-BART & $47.90\%^{*}$ & - & - & - & - & - \\
\hline
\multirow{4}{*}{\textit{GPT-3.5}}& Standard & 13.06\% & 35.58\% & 10.90\% & 6.49\% & 24.64\% & 16.51\% \\
&CoT & 33.06\% & 66.56\% & 13.27\% & 12.99\% & 46.67\% & 31.47\% \\
&Reflexion & 43.67\% & 67.64\% & 14.22\% & 15.58\% & 49.79\% & 35.28\% \\
&CoE & 52.24\% & 61.81\% & 17.06\% & 18.18\% & 49.03\% & 37.32\% \\
\hline
\multirow{4}{*}{\textit{GPT-4}}& Standard & 72.65\% & 81.13\% & 24.64\% & 25.97\% & 65.74\% & 51.10\% \\
& CoT & 76.73\% & 84.97\% & 29.86\% & 25.97\% & 69.62\% & 54.38\% \\
& Reflexion & 78.78\% & 85.12\% & 36.02\% & 27.27\% & 71.05\% & 56.49\% \\
& CoE & 76.73\% & 84.36\% & 40.28\% & 31.17\% & 71.48\% & 58.14\% \\
\hline
\multirow{4}{*}{\textit{Fine-tune}}& ORLM & 78.37\% & 84.20\% & 38.39\% & 35.06\% & 71.65\% & 59.01\% \\
& Step-Opt-Mistral-7B & 72.65\% &  82.06\% & 52.61\% & \textbf{40.26}\% & 72.15\% & 61.90\% \\
& Step-Opt-LLaMA-3-8B & \textbf{84.49}\% & \textbf{85.28}\% & \textbf{61.61}\% & 36.36\% & \textbf{77.72}\% & \textbf{66.94}\% \\
\hline
\end{tabular}
\end{center}
\vskip 0.1in
\end{table*}

\subsection{Experimental Setup}
\textbf{Dataset.} We assess our method using a range of datasets, spanning simple ones like NL4OPT \citep{nl4opt} and MAMO EasyLP \citep{mamo}, and complex ones such as MAMO ComplexLP \citep{mamo} and IndustryOR \citep{orlm}.  Answers were manually revised when needed. Examples are shown in Appendix \ref{subsec:error_example}.

\textit{NL4OPT} originates from the NeurIPS 2022 NL4Opt competition and includes 1,101 simple linear programming problems (LPs), 289 used for evaluation. We correct 16 inaccurate instances.

\textit{MAMO} contains two sub-datasets: EasyLP and ComplexLP. The former contains 652 simple LPs, and the latter 211 complex ones, all are paired with optimal solutions. We rectify 78 inaccuracies.

\textit{IndustryOR} consists of 100 complex OR problems. Many  lack essential information or accurate values, leading to 50 corrections and removal of 23 instances that fail to meet modeling criteria.

\textbf{Baselines}
\textit{tag-BART} \citep{tag-bart} is a pre-trained language model (PLM) that won 1st place in the NL4Opt competition.

\textit{Standard}, \textit{CoT} (Chain-of-Thought) \citep{cot}, and \textit{Reflexion} \citep{reflexion} represent typical prompting strategies, including direct generation, intermediate reasoning, and iterative feedback-based refinement.

\textit{Chain-of-Experts} (CoE) \citep{chain-of-expert} is a multi-agent prompting framework leveraging interactions among LLMs to enhance problem-solving.

\textit{ORLM} \citep{orlm} is a fine-tuned model using a checkpoint from Hugging Face \footnote{\url{https://huggingface.co/CardinalOperations/ORLM-LLaMA-3-8B}}, along with 3K training examples\footnote{\url{https://huggingface.co/datasets/CardinalOperations/OR-Instruct-Data-3K}}, which we also use in ablation studies.

To ensure fairness, all methods are evaluated with temperature set to 0. Fine-tuned models use greedy decoding in a zero-shot context, selecting the top-1 completion as the solution. Step-Opt and ORLM use the COPT solver to ensure alignment with the raw data format. Prompt engineering methods are evaluated using GPT-3.5 (gpt-3.5-turbo-1106) and GPT-4 (gpt-4-turbo-2024-04-09), respectively.  For additional comparison, we evaluate more advanced LLMs such as GPT-4o and Qwen2.5 \citep{qwen25} on the MAMO ComplexLP, with results provided in Appendix \ref{subsec:compare_with_4o_and_qwen25}.

\textbf{Details}
To construct the dataset, we begin with 260 examples and perform 8,400 iterations using GPT-4-turbo-0409, resulting in 4,464 examples. Further details on the instance generation can be found in Appendix~\ref{subsec:instance_generation}. We then fine-tune LLaMA-3-8B \citep{llama3} and Mistral-7B \citep{mistral} utilizing the LLaMA-Factory framework \citep{llamafactory} with the Alpaca format template \citep{alpaca}, applying the LoRA technique \citep{lora} for efficient parameter adaptation. In this setup, the input consists of a fixed prompt with a problem description, and the output includes mathematical models and the corresponding programs. Hyperparameter details are provided in Appendix \ref{subsec:hyper}.  During inference, we employ greedy search in a zero-shot context, setting the max generation length to 2,048 tokens.

\textbf{Metric.}
Considering the potential for minor discrepancies in numerical solutions, we define a comparison rule to account for small inaccuracies. Let $o$ be the output of generated programs from different methods, and $g$ denote the ground truth. The comparison is governed by the following criterion:
\begin{equation}
    \left|\frac{o - g}{g + \epsilon}\right| \leq 10^{-4},
    \label{eq:1}
\end{equation}
Where $\epsilon$ is a small number to avoid division errors; $o$ and $g$ are equivalent if they satisfy Eq.~\ref{eq:1}.

\begin{table*}[ht]
\caption{Ablation Study on different evolving methods}
\label{tab:ablation-evolution-table}
\vskip 0.1in
\begin{center}
\small 
\setlength{\tabcolsep}{4pt} 
\begin{tabular}{lcccccc}
\hline
\textbf{Method} & \textbf{NL4OPT} & \textbf{MAMO EasyLP} & \textbf{MAMO ComplexLP} & \textbf{IndustryOR} \\
\hline
Step-Opt & 77.55\% & 85.43\% & \textbf{36.02}\% & \textbf{23.38}\% \\
\hline
w/o Constraint Modification & 75.92\% & 85.58\% & 19.91\% & 15.58\% \\
w/o Objective Alteration & 77.55\% & \textbf{85.89}\% & 25.12\% & 19.48\% \\
w/o Parameter Adjustment & 73.06\% & 83.59\% & 26.07\% & 22.08\% \\ 
w/o Domain Transformation & 73.88\% & 83.13\% & 20.38\% & 18.18\% \\ 
w/o  Combination & \textbf{77.96}\% & 85.12\% & 33.65\% & 22.08\% \\ 
\hline
\end{tabular}
\end{center}
\vspace{-6pt}
\end{table*}

\begin{table*}[ht]
\caption{Comparison of Step-Opt and ORLM with 3K examples.}
\label{tab:ablation-orlm-evoor}
\vskip 0.1in
\begin{center}
\small 
\setlength{\tabcolsep}{4pt} 
\begin{tabular}{lccccccc}
\hline
\textbf{Method} & \textbf{NL4OPT} & \textbf{MAMO EasyLP} & \textbf{MAMO ComplexLP} & \textbf{IndustryOR} & \textbf{Micro Avg} & \textbf{Macro Avg} \\
\hline
Step-Opt & \textbf{78.37\%} & 84.51\% & \textbf{44.08\%} & \textbf{32.47\%} & \textbf{72.66\%} & \textbf{59.86\%} \\
\hline
ORLM & 75.92\% & \textbf{88.19\%} & 28.91\% & 25.97\% & 71.05\% & 54.75\% \\
\hline
\end{tabular}
\end{center}
\end{table*}

\begin{figure*}[ht]
\begin{center}
\centerline{\includegraphics[width=0.75\textwidth, trim=0cm 0cm 0cm 0cm, clip]{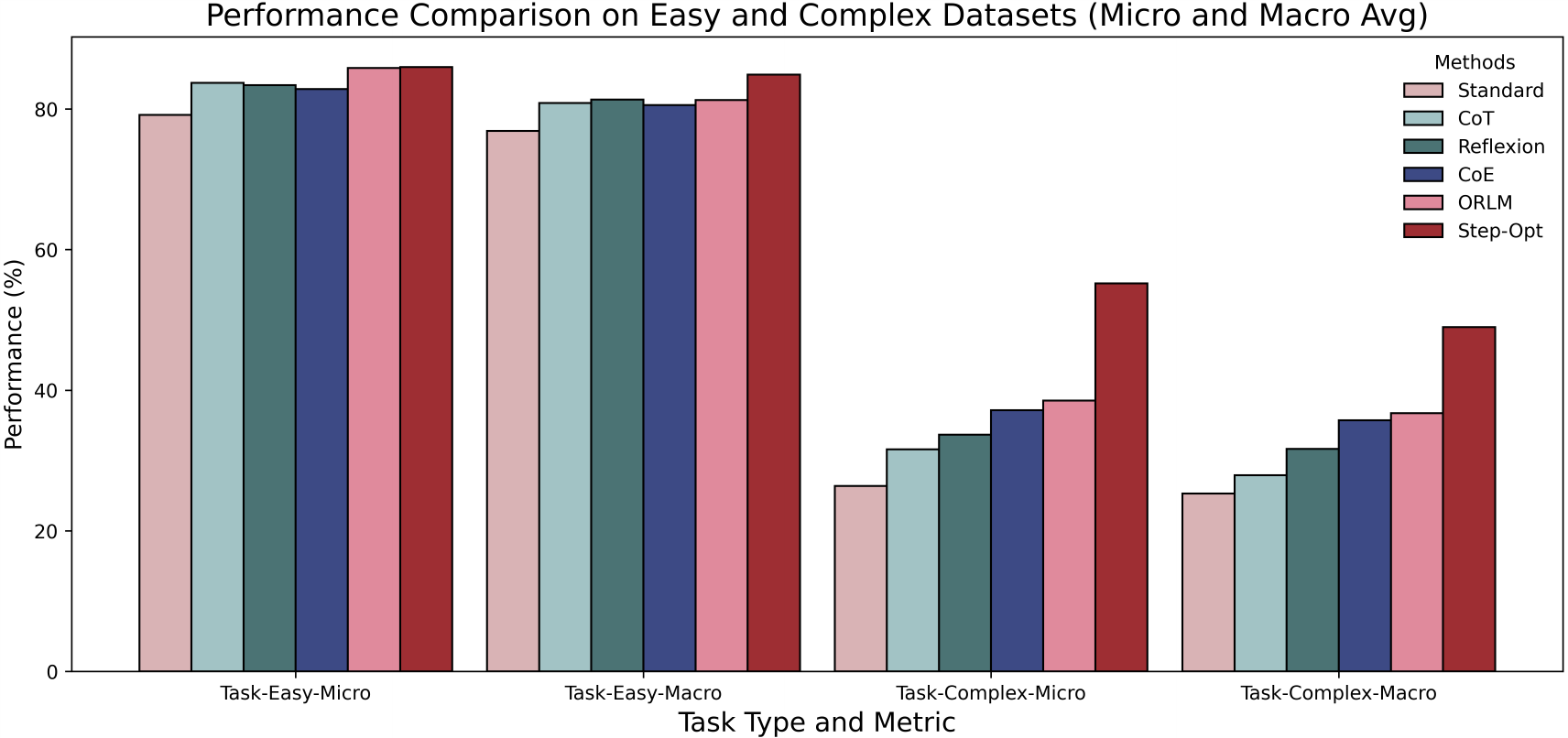}}
\caption{Performance comparison of various methods on easy and complex datasets. }
\label{fig:performance_of_easy_complex}
\end{center}
\end{figure*}

\begin{table*}[ht]
\vskip -0.2in
\caption{Comparison of Step-Opt and Step-Opt without mathematical model}
\label{tab:ablation-without_model}
\vskip 0.1in
\begin{center}
\small 
\setlength{\tabcolsep}{4pt} 
\begin{tabular}{lccccccc}
\hline
\textbf{Method} & \textbf{NL4OPT} & \textbf{MAMO EasyLP} & \textbf{MAMO ComplexLP} & \textbf{IndustryOR}  \\
\hline
Step-Opt & \textbf{84.49}\% & \textbf{85.28}\% & \textbf{61.61}\% & \textbf{36.36\%} \\
\hline 	 	 	 
Step-Opt-4.73M  & 81.22\% & 84.97\% & 50.24\% & 33.77\% \\
\hline
w/o mathematical model-4.73M & 80.00\% & 81.44\% & 45.97\% & 29.87\%  \\

\hline
\end{tabular}
\end{center}
\end{table*}

\subsection{Comparison Analysis}

As shown in Table \ref{tab:performance-table}, Step-Opts based on LLaMA-3-8B and Mistral-7B significantly outperform baselines by a large margin, including tag-BART which achieves only 47.90\% on NL4OPT despite requiring extensive manual constraint validation, lagging far behind LLM-based approaches. The best-performing Step-Opt, trained on LLaMA-3-8B, achieves state-of-the-art results on all benchmarks. This demonstrates its superior modeling capability. Notably, fine-tuned LLMs exceed the prompt engineering methods on average. However, the differences are less pronounced in the easier datasets, NL4OPT and MAMO EasyLP. The reason lies in the straightforward modeling requirements of these problems, which primarily require \deleted{the ability to understand} \added{understanding} problem descriptions—a strength of models like ChatGPT and GPT-4. In contrast, \deleted{for datasets containing more complex problems} for more complex datasets, the performance of fine-tuned models significantly improves, greatly exceeding that of prompt engineering methods. This indicates that fine-tuned models possess enhanced modeling capabilities. A prominent example is MAMO ComplexLP, where the advantage of Step-Opt-LLaMA-3-8B reaches 21.33\%.
\deleted{we further analyze the results across both simple and complex datasets. For simplicity, we select the prompt engineering method based on GPT-4 as the baseline Using GPT-4-based prompt engineering as the baselines and  Step-Opt's top model.} 

To emphasize the distinctions, we analyze results across simple and complex datasets using GPT-4 prompt engineering as the baseline compared with the top-performing Step-Opt model. As shown in Figure \ref{fig:performance_of_easy_complex}, nearly all methods perform well on simple datasets, with most achieving over 80\% accuracy, except for the Standard method. The differences between methods on simple datasets are relatively minor. In contrast, the results for complex datasets demonstrate that advanced prompt engineering techniques, such as CoE, significantly outperform Standard, CoT, and Reflexion, though they still lag behind our proposed methods. Notably, Step-Opt achieves an accuracy above 50\%, significantly surpassing existing methods and showcasing its superior modeling capabilities for complex problems. Given the intricate nature of complex problem descriptions and the advanced techniques required, our models exhibit a greater capacity to handle higher-order techniques. 


\subsection{Ablation Study}
We conduct an ablation analysis to explore the effectiveness of different evolving methods and the composition of the training data, while also facilitating a fair comparison between OR-Instruct and Step-Opt-Instruct. For all ablation experiments, we set the hyper-parameters to the same and use LLaMA-3-8B as the backbone. The parameter settings can be found in the Appendix \ref{subsec:hyper}. In addition, we further evaluate the Stepwise Validation Mechanism on the MAMO ComplexLP, the details are shown in Appendix~\ref{sec:step_ab}.

\textbf{Study on evolving methods}: Initially, we evaluate the survival rates of different generation methods, yielding the following results: 1,716 for constraint modification, 1,242 for objective alteration, 2,123 for parameter adjustment, 2,077 for domain transformation, and 455 for combination. The higher survival rates for parameter adjustment and domain transformation reflect their relative simplicity, allowing examples to pass evaluations more easily.  Conversely,  the combination is the most challenging, as it requires two sets of descriptions and solutions, often failing due to potential misalignment. The other two methods, which introduce new elements, are also more prone to errors.

Then, we randomly sample 2,000 examples without specific methods and train LLaMA-3-8B on this data. As shown in Table~\ref{tab:ablation-evolution-table},  removing domain transformation yields the worst performance, with  a clear drop across all datasets, underscoring its critical importance.  While parameter adjustment notably affects simpler benchmarks, its impact on complex datasets is limited. In contrast, both constraint modification and objective alteration exert a greater influence on complex datasets compared to easier ones. Particularly for constraint modification, it introduces additional constraints and increases the difficulty, facilitating the model's ability to process more complex conditions.

\textbf{Study on the components of training examples:} As described in Sec. \ref{label:method}, each training example includes a mathematical model and corresponding programs utilizing the COPT solver, though only the program is used for problem-solving. To assess the impact of the mathematical model, we remove this component from the entire dataset and train LLaMA-3-8B. The results, presented in Table \ref{tab:ablation-without_model}, reveal a significant performance drop upon the removal of the mathematical model. To further mitigate the influence of token count (as data without the mathematical model contain fewer tokens), we maintain a total of 4.73 million tokens across all datasets. Even with equivalent training sizes, the dataset including the mathematical model consistently outperforms the one without it. This improvement can be ascribed to the mathematical model functioning similarly to the Chain-of-Thought approach,  providing a structured framework that guides the reasoning process in a systematic manner, effectively bridging the problem description and the code solution.  In its absence, the model skips critical reasoning steps, leading to a significant reduction in performance.

\textbf{Comparison of OR-Instruct and Step-Opt-Instruct:}
ORLM gathers 686 industry cases and creates 30,000 examples with the OR-Instruct framework, including 3,000 publicly available examples. To assess the performance of OR-Instruct in comparison to Step-Opt-Instruct, we randomly select 3,000 examples for evaluation. Both datasets, each comprising 3,000 examples, are used to train LLaMA-3-8B. As illustrated in Table \ref{tab:ablation-orlm-evoor}, except for MAMO EasyLP, our method uniformly outperforms ORLM, achieving a 1.61\% improvement in micro average and a 5.11\% enhancement in macro average. The gains on more complex datasets, such as MAMO ComplexLP and IndustryOR, are even more pronounced. These advancements suggest that Step-Opt-Instruct  possesses superior capabilities and generates higher-quality data, allowing LLMs to more effectively address OR problems, particularly those of greater complexity.

\section{Conclusion}
In this paper, we present Step-Opt-Instruct, a framework that integrates iterative problem generation with a stepwise validation mechanism to enhance the capabilities of LLMs in addressing complex OR problems. By progressively increasing problem complexity and ensuring data quality through real-time validation, Step-Opt-Instruct effectively prevents error propagation by removing low-quality data during the generation process. This approach enables full automation without relying on post-processing, ensuring high-quality datasets for fine-tuning. The resulting model, Step-Opt, achieved significant performance improvements across benchmarks such as NL4OPT, MAMO, and IndustryOR, particularly excelling in complex optimization tasks. These results highlight the effectiveness of combining systematic problem generation with structured validation to significantly enhance the modeling capabilities of LLMs.

\newpage

\textbf{Limitations:} The proposed method faces difficulties in dealing with the wide variety of modeling techniques commonly used in OR, which limits its ability to handle the full range of possible scenarios. Moreover, the performance of the approach has not been fully tested across all types of OR problems. Finally, its broader application still needs to be tested in other fields to validate its applicability and adaptability.




\begin{thebibliography}{32}
\providecommand{\natexlab}[1]{#1}

\bibitem[{Achiam et~al.(2023)Achiam, Adler, Agarwal, Ahmad, Akkaya, Aleman, Almeida, Altenschmidt, Altman, Anadkat et~al.}]{gpt4}
Josh Achiam, Steven Adler, Sandhini Agarwal, Lama Ahmad, Ilge Akkaya, Florencia~Leoni Aleman, Diogo Almeida, Janko Altenschmidt, Sam Altman, Shyamal Anadkat, and 1 others. 2023.
\newblock Gpt-4 technical report.
\newblock \emph{arXiv preprint arXiv:2303.08774}.

\bibitem[{AhmadiTeshnizi et~al.(2024)AhmadiTeshnizi, Gao, and Udell}]{optimus}
Ali AhmadiTeshnizi, Wenzhi Gao, and Madeleine Udell. 2024.
\newblock Optimus: Scalable optimization modeling with (mi) lp solvers and large language models.
\newblock \emph{arXiv preprint arXiv:2402.10172}.

\bibitem[{AI@Meta(2024)}]{llama3}
AI@Meta. 2024.
\newblock \href {https://github.com/meta-llama/llama3/blob/main/MODEL_CARD.md} {Llama 3 model card}.

\bibitem[{An et~al.(2023)An, Ma, Lin, Zheng, Lou, and Chen}]{example_of_related_work6}
Shengnan An, Zexiong Ma, Zeqi Lin, Nanning Zheng, Jian-Guang Lou, and Weizhu Chen. 2023.
\newblock Learning from mistakes makes llm better reasoner.
\newblock \emph{arXiv preprint arXiv:2310.20689}.

\bibitem[{Belgacem et~al.(2020)Belgacem, Beghdad-Bey, Nacer, and Bouznad}]{or_example3}
Ali Belgacem, Kadda Beghdad-Bey, Hassina Nacer, and Sofiane Bouznad. 2020.
\newblock Efficient dynamic resource allocation method for cloud computing environment.
\newblock \emph{Cluster Computing}, 23(4):2871--2889.

\bibitem[{Bertsimas et~al.(2019)Bertsimas, Dunn, and Mundru}]{or_example1}
Dimitris Bertsimas, Jack Dunn, and Nishanth Mundru. 2019.
\newblock Optimal prescriptive trees.
\newblock \emph{INFORMS Journal on Optimization}, 1(2):164--183.

\bibitem[{Gandhi et~al.(2024)Gandhi, Gala, Viswanathan, Wu, and Neubig}]{example_of_related_work2}
Saumya Gandhi, Ritu Gala, Vijay Viswanathan, Tongshuang Wu, and Graham Neubig. 2024.
\newblock Better synthetic data by retrieving and transforming existing datasets.
\newblock \emph{arXiv preprint arXiv:2404.14361}.

\bibitem[{Ge et~al.(2022)Ge, Huangfu, Wang, Wu, and Ye}]{copt}
Dongdong Ge, Qi~Huangfu, Zizhuo Wang, Jian Wu, and Yinyu Ye. 2022.
\newblock Cardinal optimizer (copt) user guide.
\newblock \emph{arXiv preprint arXiv:2208.14314}.

\bibitem[{{Gurobi Optimization, LLC}(2024)}]{gurobi}
{Gurobi Optimization, LLC}. 2024.
\newblock \href {https://www.gurobi.com} {{Gurobi Optimizer Reference Manual}}.

\bibitem[{Hu et~al.(2021)Hu, Shen, Wallis, Allen-Zhu, Li, Wang, Wang, and Chen}]{lora}
Edward~J Hu, Yelong Shen, Phillip Wallis, Zeyuan Allen-Zhu, Yuanzhi Li, Shean Wang, Lu~Wang, and Weizhu Chen. 2021.
\newblock Lora: Low-rank adaptation of large language models.
\newblock \emph{arXiv preprint arXiv:2106.09685}.

\bibitem[{Huang et~al.(2024)Huang, Shen, Hu, Gao, and Wang}]{mamo}
Xuhan Huang, Qingning Shen, Yan Hu, Anningzhe Gao, and Benyou Wang. 2024.
\newblock Mamo: a mathematical modeling benchmark with solvers.
\newblock \emph{arXiv preprint arXiv:2405.13144}.

\bibitem[{Jiang et~al.(2023)Jiang, Sablayrolles, Mensch, Bamford, Chaplot, Casas, Bressand, Lengyel, Lample, Saulnier et~al.}]{mistral}
Albert~Q Jiang, Alexandre Sablayrolles, Arthur Mensch, Chris Bamford, Devendra~Singh Chaplot, Diego de~las Casas, Florian Bressand, Gianna Lengyel, Guillaume Lample, Lucile Saulnier, and 1 others. 2023.
\newblock Mistral 7b.
\newblock \emph{arXiv preprint arXiv:2310.06825}.

\bibitem[{Kani and Gangwar(2022)}]{tag-bart}
Nickvash Kani and Neeraj Gangwar. 2022.
\newblock Tagged input and decode all-at-once strategy.
\newblock \url{https://github.com/MLPgroup/nl4opt-generation}.

\bibitem[{Li et~al.(2023{\natexlab{a}})Li, Mellou, Zhang, Pathuri, and Menache}]{optguide}
Beibin Li, Konstantina Mellou, Bo~Zhang, Jeevan Pathuri, and Ishai Menache. 2023{\natexlab{a}}.
\newblock Large language models for supply chain optimization.
\newblock \emph{arXiv preprint arXiv:2307.03875}.

\bibitem[{Li et~al.(2023{\natexlab{b}})Li, Zhang, and Mak-Hau}]{or_data_2}
Qingyang Li, Lele Zhang, and Vicky Mak-Hau. 2023{\natexlab{b}}.
\newblock Synthesizing mixed-integer linear programming models from natural language descriptions.
\newblock \emph{arXiv preprint arXiv:2311.15271}.

\bibitem[{Oh et~al.(2023)Oh, Lee, and Jung}]{example_of_related_work3}
Seokjin Oh, Su~Ah Lee, and Woohwan Jung. 2023.
\newblock Data augmentation for neural machine translation using generative language model.
\newblock \emph{arXiv preprint arXiv:2307.16833}.

\bibitem[{Pan et~al.(2023)Pan, Cadamuro, and Groh}]{example_of_related_work4}
Yan Pan, Davide Cadamuro, and Georg Groh. 2023.
\newblock Data-augmented task-oriented dialogue response generation with domain adaptation.
\newblock In \emph{Proceedings of the 37th Pacific Asia Conference on Language, Information and Computation}, pages 96--106.

\bibitem[{Pereira et~al.(2022)Pereira, Oliver, Francisco, Cunha~Jr, and Gomes}]{or_example2}
Jo{\~a}o Luiz~Junho Pereira, Guilherme~Ant{\^o}nio Oliver, Matheus~Brendon Francisco, Sebastiao~Simoes Cunha~Jr, and Guilherme~Ferreira Gomes. 2022.
\newblock A review of multi-objective optimization: methods and algorithms in mechanical engineering problems.
\newblock \emph{Archives of Computational Methods in Engineering}, 29(4):2285--2308.

\bibitem[{Prasath and Karande(2023)}]{or_data_1}
Ganesh Prasath and Shirish Karande. 2023.
\newblock Synthesis of mathematical programs from natural language specifications.
\newblock \emph{arXiv preprint arXiv:2304.03287}.

\bibitem[{Ramamonjison et~al.(2023)Ramamonjison, Yu, Li, Li, Carenini, Ghaddar, He, Mostajabdaveh, Banitalebi-Dehkordi, Zhou et~al.}]{nl4opt}
Rindranirina Ramamonjison, Timothy Yu, Raymond Li, Haley Li, Giuseppe Carenini, Bissan Ghaddar, Shiqi He, Mahdi Mostajabdaveh, Amin Banitalebi-Dehkordi, Zirui Zhou, and 1 others. 2023.
\newblock Nl4opt competition: Formulating optimization problems based on their natural language descriptions.
\newblock In \emph{NeurIPS 2022 Competition Track}, pages 189--203. PMLR.

\bibitem[{Shinn et~al.(2024)Shinn, Cassano, Gopinath, Narasimhan, and Yao}]{reflexion}
Noah Shinn, Federico Cassano, Ashwin Gopinath, Karthik Narasimhan, and Shunyu Yao. 2024.
\newblock Reflexion: Language agents with verbal reinforcement learning.
\newblock \emph{Advances in Neural Information Processing Systems}, 36.

\bibitem[{Tang et~al.(2024)Tang, Huang, Zheng, Hu, Wang, Ge, and Wang}]{orlm}
Zhengyang Tang, Chenyu Huang, Xin Zheng, Shixi Hu, Zizhuo Wang, Dongdong Ge, and Benyou Wang. 2024.
\newblock Orlm: Training large language models for optimization modeling.
\newblock \emph{arXiv preprint arXiv:2405.17743}.

\bibitem[{Taori et~al.(2023)Taori, Gulrajani, Zhang, Dubois, Li, Guestrin, Liang, and Hashimoto}]{alpaca}
Rohan Taori, Ishaan Gulrajani, Tianyi Zhang, Yann Dubois, Xuechen Li, Carlos Guestrin, Percy Liang, and Tatsunori~B Hashimoto. 2023.
\newblock Stanford alpaca: An instruction-following llama model.

\bibitem[{Wang et~al.(2022)Wang, Kordi, Mishra, Liu, Smith, Khashabi, and Hajishirzi}]{self-instruct}
Yizhong Wang, Yeganeh Kordi, Swaroop Mishra, Alisa Liu, Noah~A Smith, Daniel Khashabi, and Hannaneh Hajishirzi. 2022.
\newblock Self-instruct: Aligning language models with self-generated instructions.
\newblock \emph{arXiv preprint arXiv:2212.10560}.

\bibitem[{Wei et~al.(2022)Wei, Wang, Schuurmans, Bosma, Xia, Chi, Le, Zhou et~al.}]{cot}
Jason Wei, Xuezhi Wang, Dale Schuurmans, Maarten Bosma, Fei Xia, Ed~Chi, Quoc~V Le, Denny Zhou, and 1 others. 2022.
\newblock Chain-of-thought prompting elicits reasoning in large language models.
\newblock \emph{Advances in neural information processing systems}, 35:24824--24837.

\bibitem[{Wei et~al.(2024)Wei, Wang, Liu, Ding, and Zhang}]{example_of_related_work1}
Yuxiang Wei, Zhe Wang, Jiawei Liu, Yifeng Ding, and Lingming Zhang. 2024.
\newblock Magicoder: Empowering code generation with oss-instruct.
\newblock In \emph{Forty-first International Conference on Machine Learning}.

\bibitem[{Xiao et~al.(2023)Xiao, Zhang, Wu, Xu, Wang, Han, Fu, Zhong, Zeng, Song et~al.}]{chain-of-expert}
Ziyang Xiao, Dongxiang Zhang, Yangjun Wu, Lilin Xu, Yuan~Jessica Wang, Xiongwei Han, Xiaojin Fu, Tao Zhong, Jia Zeng, Mingli Song, and 1 others. 2023.
\newblock Chain-of-experts: When llms meet complex operations research problems.
\newblock In \emph{The Twelfth International Conference on Learning Representations}.

\bibitem[{Xu et~al.(2024)Xu, Sun, Zheng, Geng, Zhao, Feng, Tao, Lin, and Jiang}]{wizardlm}
Can Xu, Qingfeng Sun, Kai Zheng, Xiubo Geng, Pu~Zhao, Jiazhan Feng, Chongyang Tao, Qingwei Lin, and Daxin Jiang. 2024.
\newblock Wizardlm: Empowering large pre-trained language models to follow complex instructions.
\newblock In \emph{The Twelfth International Conference on Learning Representations}.

\bibitem[{Yang et~al.(2024{\natexlab{a}})Yang, Yang, Zhang, Hui, Zheng, Yu, Li, Liu, Huang, Wei et~al.}]{qwen25}
An~Yang, Baosong Yang, Beichen Zhang, Binyuan Hui, Bo~Zheng, Bowen Yu, Chengyuan Li, Dayiheng Liu, Fei Huang, Haoran Wei, and 1 others. 2024{\natexlab{a}}.
\newblock Qwen2. 5 technical report.
\newblock \emph{arXiv preprint arXiv:2412.15115}.

\bibitem[{Yang et~al.(2024{\natexlab{b}})Yang, Huang, Shi, Feng, Song, Wang, Liang, and Tang}]{or_data_3}
Zhicheng Yang, Yinya Huang, Wei Shi, Liang Feng, Linqi Song, Yiwei Wang, Xiaodan Liang, and Jing Tang. 2024{\natexlab{b}}.
\newblock Benchmarking llms for optimization modeling and enhancing reasoning via reverse socratic synthesis.
\newblock \emph{arXiv preprint arXiv:2407.09887}.

\bibitem[{Zheng et~al.(2024)Zheng, Zhang, Zhang, Ye, Luo, Feng, and Ma}]{llamafactory}
Yaowei Zheng, Richong Zhang, Junhao Zhang, Yanhan Ye, Zheyan Luo, Zhangchi Feng, and Yongqiang Ma. 2024.
\newblock \href {http://arxiv.org/abs/2403.13372} {Llamafactory: Unified efficient fine-tuning of 100+ language models}.
\newblock In \emph{Proceedings of the 62nd Annual Meeting of the Association for Computational Linguistics (Volume 3: System Demonstrations)}, Bangkok, Thailand. Association for Computational Linguistics.

\bibitem[{Zhou et~al.(2024)Zhou, Zhang, Wang, Chen, Zhao, Sha, Sheng, Wang, and Wen}]{example_of_related_work5}
Kun Zhou, Beichen Zhang, Jiapeng Wang, Zhipeng Chen, Wayne~Xin Zhao, Jing Sha, Zhichao Sheng, Shijin Wang, and Ji-Rong Wen. 2024.
\newblock Jiuzhang3. 0: Efficiently improving mathematical reasoning by training small data synthesis models.
\newblock \emph{arXiv preprint arXiv:2405.14365}.

\end{thebibliography}

\appendix

\newpage
\appendix

\section{Appendix}\label{sec:appendix}
\subsection{Example for training data} \label{subsec:training example}
\begin{figure*}[htbp]
\begin{center}
\includegraphics[width=1\textwidth, trim=0cm 0cm 0cm 0cm, clip]{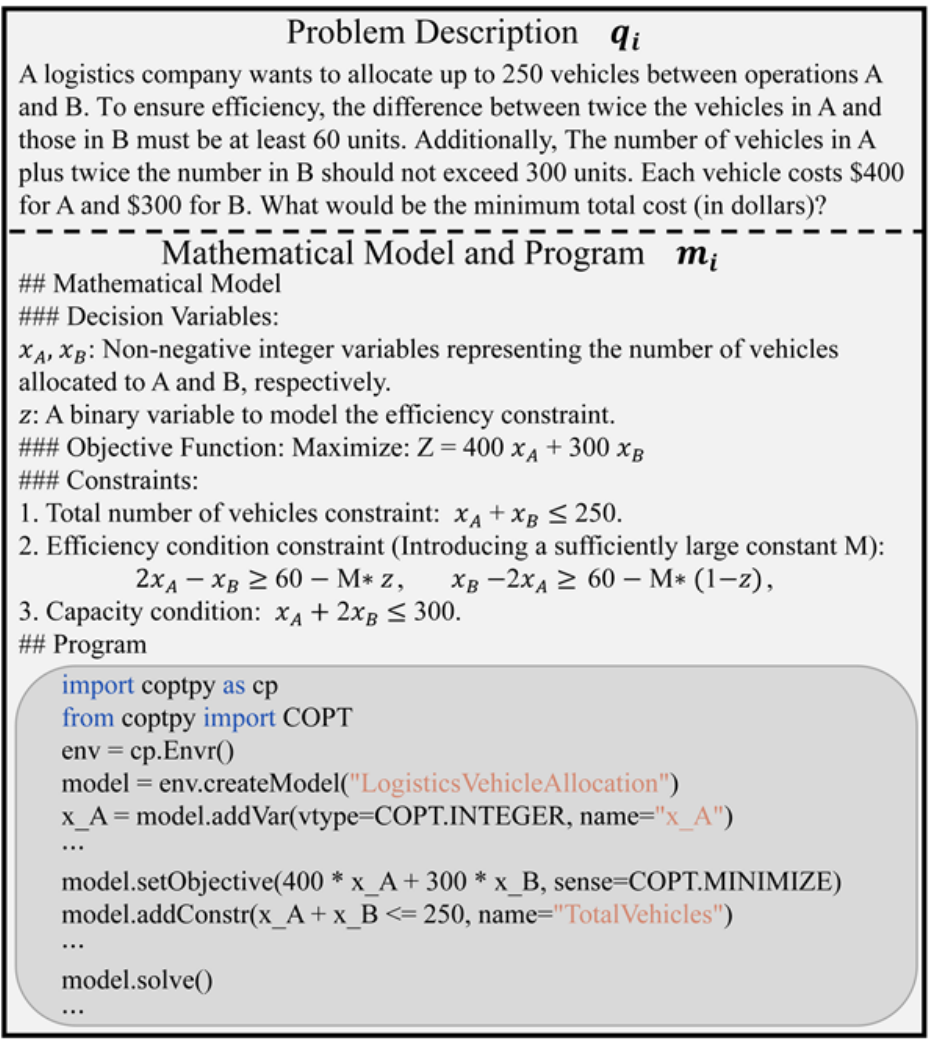}
\end{center}
\caption{Examples of training data.}
\label{fig:train_data}
\end{figure*}
We use COPT \citep{copt} as the default solver in our experiments.

\subsection{Examples for modifications of test sets}\label{subsec:error_example}
\textbf{NL4OPT, Entry \#228} : Wrong variable definition
\par Problem: A macro-counting fitness guru only eats salmon and eggs. Each bowl of salmon contains 300 calories, 15 grams of protein, and 80 mg of sodium. Each bowl of eggs contains 200 calories, 8 grams of protein, and 20 mg of sodium. Since the fitness guru has a limit to how many eggs he would like to eat, at most 40\% of his meals can be eggs. The fitness guru needs to eat at least 2000 calories and 90 grams of protein. How many of each type of meal should he eat to minimize his sodium intake? 
\textbf{Answer:} 430.7692307692307
\par
The answer is initially derived by treating the number of salmon and egg bowls as continuous variables. However, since the number of bowls should be integers, the correct solution is adjusted, and the actual answer is 460.
\\ 

\textbf{MAMO EasyLP, Entry \#216} : Incorrect Handling of Absolute Value Constraint

\par Problem: A retail manager is planning to allocate resources across three different departments: purchasing (X), sales (Y), and logistics (Z). These departments have different cost per unit of resource allocated, with \$5 for X, \$3 for Y, and \$4 for Z. The objective is to minimize the total cost while meeting certain operational constraints. The combined resources allocated to purchasing and sales cannot exceed 1000 units due to budget limitations. Similarly, the combined resources allocated to sales and logistics cannot exceed 800 units due to manpower availability. To ensure a balanced operation, the difference in resource allocation between purchasing and logistics should be at least 200 units. Given that each department has specific bounds on resource allocation (Purchasing can have up to 500 units, Sales up to 300 units, Logistics up to 200 units) and that allocations must be whole numbers due to indivisible nature of the resources being allocated:What is the minimum total cost required for this scenario? 
type of meal should he eat to minimize his sodium intake?
\textbf{Answer:} 1000
\par
The initial solution was derived without successfully establishing an absolute value constraint for "the difference in resource allocation between purchasing and logistics should be at least 200 units." Instead, only the constraint for one side (greater than or equal to 200) is retained, leading to an error. That is "model.addConstr(x - z >= 200, name=ResourceDifferenceConstraint)" in the program. The correct solution, considering both sides of the absolute value constraint, yields an actual cost of 800.

\textbf{MAMO ComplexLP, Entry \#216} : Incorrect Handling of  Subtour Elimination

\par Problem: Imagine a logistics manager tasked with planning a delivery route for a truck that needs to visit four different cities to distribute goods. The cities are identified numerically as 1, 2, 3, and 4. The truck can start its journey from any of these cities but must travel to each city exactly once and then return to the starting point. The objective is to arrange this route in such a way that the total travel cost is minimized. The costs associated with traveling between the cities are as follows:  The cost to travel from City 1 to City 2 is 52 units, to City 3 is 89 units, and to City 4 is 11 units. From City 2, it costs 52 units to reach City 1, 14 units to get to City 3, and 13 units to City 4. Traveling from City 3, the costs are 89 units to City 1, 14 units to City 2, and 87 units to City 4. Lastly, from City 4, it costs 11 units to go to City 1, 13 units to City 2, and 87 units to City 3. What is the minimum total travel cost for the truck to visit each city exactly once and return to the starting city?
\textbf{Answer:} 50
\par
The initial solution was derived without successfully establishing the subtour elimination constraint for the Traveling Salesman Problem (TSP). As a result, subtours were not eliminated properly, leading to an incorrect minimum total travel cost of 50 units. The correct solution, ensuring that subtours are eliminated and all cities are visited exactly once, yields an actual minimum total travel cost of 127 units.

\textbf{IndustryOR, Entry \#86}: Missing Number 

Problem: Fighter jets are important combat tools, but in order for them to be effective, 
there must be enough pilots. Therefore, in addition to a portion of the produced fighter jets being used directly for combat, another portion needs to be allocated for pilot training. It is known that the number of fighter jets produced each year is $a_j (j=1,\cdots,n)$, and each fighter jet can train k pilots per year. How should the production of fighter jets be allocated each year to maximize their contribution to national defense over a period of n year?
There is no numerical value for all parameters.

\newpage

\subsection{Prompt Templates for Complexity-Evolving}\label{subsec:depth}

\begin{figure}[htbp]
\newlength{\oldcolumnsep}
\setlength{\oldcolumnsep}{\columnsep}
\setlength{\columnsep}{0pt}

\noindent 
\begin{minipage}{1\textwidth}
\begin{tcolorbox}[colback=blue!12!white,colframe=blue!75!black,
title=Prompt for objective alteration of Complexity-Evolving,
    left=1mm,  
    right=1mm, 
    top=1mm,   
    bottom=1mm 
]

Assume you are an expert in combinatorial optimization modeling. Modify the objective function to either transform the current objective into a different metric or add a new objective to convert it into a multi-objective optimization problem, while retaining its logical structure. \textcolor[RGB]{192,0,0}{The modifications or additions to the objective function should be substantial and not merely changes to coefficients. If there are already two or more objective functions, no new objectives may be added; only the existing objectives can be modified.} The newly generated problems should align with real-world scenarios.
\textbf{You only need to produce a single new problem and do not solve it.}

\textbf{Given example1:}                             \{Here is Example1\}

\textbf{Given example2:}                             \{Here is Example2\}

\textbf{Given input:}                    \{Here is the original problem description\}

\textbf{Answer:}

\vspace{0.0mm}
\end{tcolorbox}
\end{minipage}
\end{figure}

\begin{figure}[htbp]
\setlength{\oldcolumnsep}{\columnsep}
\setlength{\columnsep}{0pt}

\noindent 
\begin{minipage}{1\textwidth}
\begin{tcolorbox}[colback=blue!12!white,colframe=blue!75!black,
title=Prompt for parameter adjustment of Complexity-Evolving,
    left=1mm,  
    right=1mm, 
    top=1mm,   
    bottom=1mm 
]

Assume you are an expert in combinatorial optimization modeling. Adjust the parameters of the given problem while retaining its logical structure, constraints, and objective. \textcolor[RGB]{192,0,0}{When introducing a new entity, restrict the introduction to at most one new entity to control the complexity of the problem.} The newly generated problems should align with real-world scenarios.
\textbf{You only need to produce a single new problem and do not solve it.}

\textbf{Given example1:}                             \{Here is Example1\}

\textbf{Given example2:}                             \{Here is Example2\}

\textbf{Given input:}                    \{Here is the original problem description\}

\textbf{Answer:}
\vspace{0.0mm}
\end{tcolorbox}
\end{minipage}
\end{figure}

\begin{figure}[htbp]
\setlength{\oldcolumnsep}{\columnsep}
\setlength{\columnsep}{0pt}

\noindent 
\begin{minipage}{1\textwidth}
\begin{tcolorbox}[colback=blue!12!white,colframe=blue!75!black,
title=Prompt for constraint modification of Complexity-Evolving,
    left=1mm,  
    right=1mm, 
    top=1mm,   
    bottom=1mm 
]

Assume you are an expert in combinatorial optimization modeling. 
Modify constraints or add new constraints based on the given problem while retaining its logical structure. \textcolor[RGB]{192,0,0}{Note that the modifications or additions to the constraints should be limited to a maximum of one.} The newly generated problems should align with real-world scenarios.
\textbf{You only need to produce a single new problem and do not solve it.}

\textbf{Given example1:}                             \{Here is Example1\}

\textbf{Given example2:}                             \{Here is Example2\}

\textbf{Given input:}                    \{Here is the original problem description\}

\textbf{Answer:}

\vspace{0.0mm}
\end{tcolorbox}

\end{minipage}
\end{figure}

\newpage
~
\newpage
\subsection{Prompt Templates for Scope-Evolving} \label{subsec:breadth}

\begin{figure}[htbp]
\setlength{\oldcolumnsep}{\columnsep}
\setlength{\columnsep}{0pt}

\noindent 
\begin{minipage}{1\textwidth}
\begin{tcolorbox}[colback=blue!12!white,colframe=blue!75!black,
title=Prompt for domain transformation of Scope-Evolving,
    left=1mm,  
    right=1mm, 
    top=1mm,   
    bottom=1mm 
]

Assume you are an expert in combinatorial optimization modeling. 
Transform the basic structure of the given problem into a different application domain while retaining its logical structure and constraints. The new application domain can include, but is not limited to, the following: the following: Education, Manufacturing, Logistics, Retail, Agriculture, IT Services, Healthcare, Event Planning, Construction, Entertainment, Research and Development, Hospitality, Defense, Energy Sector, Transportation, and Telecommunications.
\textbf{You only need to produce a single new problem and do not solve it.}

\textbf{Given example1:}                             \{Here is Example1\}

\textbf{Given example2:}                             \{Here is Example2\}

\textbf{Given input:}                    \{Here is the original problem description\}

\textbf{Answer:}

\vspace{0.0mm}
\end{tcolorbox}

\end{minipage}
\end{figure}

\begin{figure}[htbp]
\setlength{\oldcolumnsep}{\columnsep}
\setlength{\columnsep}{0pt}

\noindent 
\begin{minipage}{1\textwidth}
\begin{tcolorbox}[colback=blue!12!white,colframe=blue!75!black,
title=Prompt for combination of Scope-Evolving,
    left=1mm,  
    right=1mm, 
    top=1mm,   
    bottom=1mm 
]

Assume you are an expert in combinatorial optimization modeling. Given two problems (\# Problem1 and \# Problem2), generate a new problem. The new problem should be similar in length to one of the original problems but should belong to a different domain and have distinct specific details. The newly generated problem should align with real-world scenarios. \textbf{You only need to produce a single new problem and do not solve it.}

\textbf{Given example1:}                             \{Here is Example1\}

\textbf{Given example2:}                             \{Here is Example2\}

\textbf{Given input:}                    

\textbf{\# Problem1:} \{Here is the first problem description\}

\textbf{\# Problem2:} \{Here is the second problem description\}

\textbf{Answer:}

\vspace{0.0mm}
\end{tcolorbox}

\end{minipage}
\end{figure}

\subsection{Prompt Templates for checkers and regeneration}\label{others}

\begin{figure}[htbp]
\setlength{\oldcolumnsep}{\columnsep}
\setlength{\columnsep}{0pt}

\noindent 
\begin{minipage}{1\textwidth}
\begin{tcolorbox}[colback=blue!12!white,colframe=blue!75!black,
title=Prompt for regenerating the solution,
    left=1mm,  
    right=1mm, 
    top=1mm,   
    bottom=1mm 
]

\# Solution is the mathematic model and program of \# Problem. An \'Error\' was detected in \# Solution.  Please regenerate the solution based on the \'Error\'. Ensure that the new solution correctly addresses the problem while maintaining the same format and structure as the original \# Solution, with only the necessary corrections and improvements. No additional explanations are required.

$\#$ Problem:                             \{Here is the generated problem description\}

$\#$ Solution:                              \{Here is the mathematic model and program for $\#$ Problem\}

\'Error\':                              \{Here is the error\}

\textbf{Given Example1:}             \{Here is Example1\}

\textbf{Given Example2:}             \{Here is Example2\}

\textbf{Answer:}

\vspace{0.0mm}
\end{tcolorbox}

\end{minipage}
\end{figure}

\newpage
~\newpage

\begin{figure}[htbp]
\setlength{\oldcolumnsep}{\columnsep}
\setlength{\columnsep}{0pt}

\noindent 
\begin{minipage}{1\textwidth}
\begin{tcolorbox}[colback=blue!12!white,colframe=blue!75!black,
title=$\#$ Task Description: Comprehensive Constraint Validation for OR Problems.
,
    left=1mm,  
    right=1mm, 
    top=1mm,   
    bottom=1mm 
]
\textbf{Important: The checks must be based on the problem description and common sense. No assumptions or conjectures should be made.}

\textbf{\#\# Solution Description:} To verify the correctness of all constraints in the \texttt{"\#\# Mathematical Model"} for a problem, follow this structured approach:

\#\#\# Step 1: Extract Constraint Definitions

1. In the \texttt{"\#\# Mathematical Model"}, identify constraints under \texttt{"\#\#\# Constraints"}.

2. In the \texttt{"\#\# Python Code Solution Using coptpy"}, find where \texttt{model.addConstr} is used.

\#\#\# Step 2: Validate Constraint Alignment with Problem Objectives

\qquad \qquad \qquad \qquad \qquad \qquad $\cdots$

\#\#\# Step 3: Special Checks on Big-M Method Applications

1. \textbf{Absolute Value Constraints:} For constraints \( |x_i - x_j| \geq a \)), verify the use of the Big-M method:

  Introduce a binary decision variable \( y \) for each constraint, and a sufficiently large constant \( M \).  
  Split into two constraints:
  $x_i - x_j \geq a - M \cdot y $ and $ x_j - x_i \geq a - M \cdot (1 - y)$

  2. \textbf{K-Way Selection Constraints:} For "at most \( K \)" selections from \( N \) types, confirm constraints:
  $
  \sum_{i=1}^{N} y_i \leq K$ and $x_i \leq M \cdot y_i
  $
  where \( y_i \) is a binary variable and \( M \) is a sufficiently large constant.

\textbf{\#\#\# Step 4: Confirm Consistency with Python Code}

Ensure that the constraints defined in the mathematical model are accurately translated into the code.

\textbf{If no errors are found:} ``There are no errors found.'' \\
\textbf{If errors are identified:} Output ``ERROR:'' followed by the issue and advice for correction.

\vspace{0.0mm}
\end{tcolorbox}

\end{minipage}
\end{figure}

\begin{figure}[htbp]
\setlength{\oldcolumnsep}{\columnsep}
\setlength{\columnsep}{0pt}

\noindent 
\begin{minipage}{1\textwidth}
\begin{tcolorbox}[colback=blue!12!white,colframe=blue!75!black,
title=Prompt for regenerating the problem description,
    left=1mm,  
    right=1mm, 
    top=1mm,   
    bottom=1mm 
]

The \# Problem is a generated problem but has some \'Error\'. Please regenerate the problem description based on the \'Error\'. Ensure that the new problem follows the same format and structure as \# Problem, with only the necessary corrections and detail enhancements. \textbf{No solution or any other additional explanations are required.}

$\#$ Problem:                             \{Here is the generated problem description\}

\'Error\':                              \{Here is the error\}

\textbf{Example1:}                    

$\#$ Problem:                             \{example problem 1\}

\'Error\':                              \{example error 1\}

\'Regenerate\':       \{example regenerate 1\}

\textbf{Example2:}                    

$\#$ Problem:                             \{example problem 2\}

\'Error\':                              \{example error 2\}

\'Regenerate\':       \{example regenerate 2\}

\textbf{Answer:}

\vspace{0.0mm}
\end{tcolorbox}

\end{minipage}
\end{figure}


\newpage
~
\newpage

\begin{table*}[ht]
\vspace{-2pt}
\caption{Comparison of Stepwise Validation Mechanism and Other Prompt Engineering Methods on MAMO ComplexLP}
\label{tab:ab_of_step}
\begin{center}
\small 
\setlength{\tabcolsep}{4pt} 
\begin{tabular}{lccc}
\hline
\textbf{Method\textbackslash Model} & \textbf{GPT-3.5} & \textbf{GPT-4} & \textbf{GPT-4o}   \\
\hline
Standard & 10.90\% & 24.64\% & 46.92\% \\
\hline 	 	 	 
CoT  & 13.27\% & 29.86\% & 49.29\% \\
\hline
Reflexion  & 14.22\% & 36.02\% & 48.34\% \\
\hline
CoE  & \textbf{17.06\%} & 40.28\% & \textbf{54.03\%} \\
\hline
 Stepwise Validation Mechanism  & 16.59\% & \textbf{42.18\%} & \textbf{50.71\%} \\

\hline
\end{tabular}
\end{center}
\end{table*}

\begin{table*}[ht]
\caption{Performance Comparison of Various Methods on MAMO ComplexLP}
\label{tab:performance-on-mamo_complexlp}
\begin{center}
\small 
\setlength{\tabcolsep}{4pt} 
\begin{tabular}{lccccc}
\hline
\textbf{Method} & \textbf{Standard}	&\textbf{CoT}	&\textbf{Reflexion}	& \textbf{CoE}	&\textbf{Fine-Tuning} \\

\hline
GPT-3.5 & 10.90\% & 13.27\% & 14.22\% & 17.06\% & - \\
GPT-4	&24.64\%	&29.86\%	&36.02\%	&40.28\% & -\\
GPT-4o	&46.92\%	&49.29\%	&48.34\%	&54.03\% & -\\
Qwen2.5-72B-Instruct	&46.45\%	&45.97\%	&47.87\%	&51.66\% & -\\
ORLM &  - & - & - & - & 38.39\%\\
Step-Opt-Mistral-7B&  - & - & - & - & 52.61\%\\
Step-Opt-LLaMA-3-8B &  - & - & - & - & \textbf{61.61\%}\\
\hline
\end{tabular}
\end{center}
\end{table*}

\begin{table*}[htbp]
\centering
\caption{Hyper-parameters for \textbf{ablation experiments}.}
\begin{tabular}{lcccc}
\hline
 \textbf{BatchSize Per GPU} &  \textbf{Gradient Accumulation}&\textbf{Learning rate} & \textbf{Epochs} \\ \hline
 4 & 8 & 1.25$\times 10^{-4}$ &10 \\ \hline
\end{tabular}
\end{table*}

\newpage
~
\subsection{Performance Comparison on MAMO ComplexLP}\label{subsec:compare_with_4o_and_qwen25}

To further validate our results, we conducted comparative studies on the MAMO ComplexLP dataset, involving leading proprietary LLM GPT-4o-2024-08-06 and an advanced open-source LLM Qwen2.5-72B-Instruct. These comparisons provide additional context to the effectiveness of our Step-Opt framework. As shown in Table~\ref{tab:performance-on-mamo_complexlp}, proprietary models like GPT-4o demonstrate notable improvements, achieving a maximum accuracy of 54.03\% with CoE and consistently outperforming earlier versions like GPT-4 and GPT-3.5. Similarly, open-source models such as Qwen2.5 achieve competitive results, with Reflexion reaching 47.87\% and CoE achieving 51.66\%. These findings indicate that open-source models are steadily narrowing the gap with proprietary counterparts, even without task-specific fine-tuning. 

Despite these advancements, Step-Opt still demonstrates significant superiority, achieving the highest accuracy of 61.61\% with Step-Opt-LLaMA-3-8B, surpassing GPT-4o and other baselines. Step-Opt-Mistral-7B also achieves 52.61\%, further showcasing the effectiveness of our framework. These results emphasize the impact of Step-Opt’s task-specific training data in elevating model performance across diverse problem formulations. 

By generating high-quality, diverse datasets, Step-Opt addresses a key challenge in structured optimization tasks: enabling LLMs to better handle complex problems. The consistent performance of Step-Opt-trained models highlights the importance of integrating precise, task-specific data into fine-tuning pipelines, paving the way for more reliable and effective solutions to real-world optimization challenges.

\subsection{Additional Details on Instance Generation}
\label{subsec:instance_generation}

The instance generation involved 64K queries, and the number of tokens was 179M. On average, each generation iteration required approximately 7.66 queries, with 3.14 queries dedicated to generating and validating the problem description, and 4.52 queries used for solution generation and validation. Of the total tokens, 39M were allocated to generating and validating the problem description, while the remaining 140M were used for solution generation and validation. Additionally, 8,400 generations were conducted, yielding 4,464 samples, indicating that 46.86\% of the generated samples were discarded.

\subsection{Hyper-parameters for Training Step-Opt and baselines}\label{subsec:hyper}

\begin{table*}[h!]
\centering
\caption{Hyper-parameters for Training Step-Opts.}
\begin{tabular}{lcccc}
\hline
\textbf{Backbone} & \textbf{BatchSize Per GPU} &  \textbf{Gradient Accumulation}&\textbf{Learning rate} & \textbf{Epochs} \\ \hline
Mistral-7B & 4 & 8 & 1.25$\times 10^{-4}$ &10 \\ \hline
LLaMA-3-8B & 4 & 8 & 1.25$\times 10^{-4}$ &12 \\ \hline
\end{tabular}
\end{table*}

All experiments are conducted on a single GPU server equipped with eight A100 GPUs, each with 40GB of memory. In experiment, we report the best results of all checkpoints. The maximum token is limited to 2,500. The hyper-parameters for training Step-Opts are as follows:

\subsection{Study on Stepwise Validation Mechanism}\label{sec:step_ab}
To evaluate the impact of the proposed Stepwise Validation Mechanism, we conducted experiments on the MAMO Complex dataset. The results are summarized in Table~\ref{tab:ab_of_step}. The results demonstrate that the Stepwise Validation Mechanism delivers consistent improvements across different GPT models when compared to Standard, CoT, and Reflexion methods. For GPT-4, our framework achieves the highest accuracy (42.18\%), outperforming all other methods. However, CoE remains superior for GPT-3.5 and GPT-4o, reflecting the strength of its iterative reflection mechanism in these cases.

In contrast, the Stepwise Validation Mechanism emphasizes real-time validation and correction during the modeling process, , avoiding the additional complexity of reflection. This streamlined approach proves particularly effective for LLMs, as demonstrated by its superior performance with GPT-4. Although CoE excels in certain cases, our method offers a robust and efficient alternative.

Additionally, It is important to consider the inherent difficulty of solving tasks directly during testing, as all methods must generate solutions from scratch. However, when used for data generation, the Stepwise Validation Mechanism can reference the solution of the original problem to generate solutions for new problems. By focusing only on the newly added or modified components, the mechanism significantly reduces the modeling difficulty. This advantage is not available during testing, where tasks must be solved entirely independently, but it underscores the potential of Stepwise Validation Mechanism for facilitating high-quality data generation.



\end{document}